\documentclass[conference]{IEEEtran}
\IEEEoverridecommandlockouts
\usepackage{iftex}
\ifLuaTeX
  \usepackage{graphicx}
\else\ifPDFTeX
  \usepackage{graphicx}
\else
  \usepackage[dvipdfmx]{graphicx}
\fi\fi

\usepackage{amsmath}
\usepackage{amsfonts}
\usepackage{amsbsy}
\usepackage{bm}

\ifLuaTeX
  \usepackage{newtxtext}
  \usepackage{newtxmath}
\fi

\usepackage{textcomp}
\usepackage{xcolor}
\def\BibTeX{{\rm B\kern-.05em{\sc i\kern-.025em b}\kern-.08em
    T\kern-.1667em\lower.7ex\hbox{E}\kern-.125emX}}

\usepackage{stmaryrd}
\usepackage{ascmac}
\usepackage{cite}
\usepackage{amsbsy}
\usepackage{latexsym}
\usepackage[%
  caption = false,%
  font = footnotesize%
]{subfig}

\graphicspath{{figs/}}

\usepackage{mathtools}
\usepackage{enumitem}
\usepackage{mathrsfs}

\usepackage{algorithm}
\usepackage{algorithmicx, algpseudocode}

\newtheorem{proposition}{Proposition}

\newtheorem{corollary}{Corollary}

\newtheorem{remark}{Remark}

\newcommand{\dom}{\mathop{\rm dom}\nolimits}
\newcommand{\Prox}{\mathop{\rm Prox}\nolimits}

\newcommand{\Int}{\mathop{\rm int}\nolimits}
\newcommand{\soft}{\mathop{\rm soft}\nolimits}

\newcommand{\firm}{\mathop{\rm firm}\nolimits}

\newcommand{\sign}{\mathop{\rm sgn}\nolimits}

\newcommand{\lev}{\mathop{\rm lev}\nolimits}

\newcommand{\argmin}{\operatornamewithlimits{argmin}}

\DeclareMathOperator{\Poisson}{Poisson}

\begin{document}

\title{External Division of Two Bregman Proximity Operators for Poisson Inverse Problems%
  \thanks{This work was supported by the Grants-in-Aid for Scientific Research (KAKENHI) under Grant Number
    25K24422.} }

\author{\IEEEauthorblockN{Kazuki Haishima, Kyohei Suzuki, and Konstantinos Slavakis}
\IEEEauthorblockA{%
  \textit{Institute of Science Tokyo, Department of Information and Communications}\\
  Emails: \{haishima.k.7e90, suzuki.k.439f\}@m.isct.ac.jp, slavakis@ict.eng.isct.ac.jp%
}}

\maketitle

\begin{abstract}
  This paper presents a novel method for recovering sparse vectors from linear models corrupted by
  Poisson noise. The contribution is twofold. First, an operator defined via the external division of two
  Bregman proximity operators is introduced to promote sparse solutions while mitigating the estimation
  bias induced by classical $\ell_1$-norm regularization. This operator is then embedded into the already
  established NoLips algorithm, replacing the standard Bregman proximity operator in a plug-and-play
  manner. Second, the geometric structure of the proposed external-division operator is elucidated
  through two complementary reformulations, which provide clear interpretations in terms of the primal
  and dual spaces of the Poisson inverse problem. Numerical tests show that the proposed method exhibits
  more stable convergence behavior than conventional Kullback-Leibler (KL)-based approaches and achieves
  significantly superior performance on synthetic data and an image restoration problem.
\end{abstract}

\begin{IEEEkeywords}
    Inverse problems, Poisson noise, sparse modeling, Bregman divergence, proximity operator.
\end{IEEEkeywords}

\section{Introduction}

Inverse problems involving discrete event counts, such as positron emission tomography (PET) imaging and
photon counting, are typically corrupted by Poisson noise rather than Gaussian~\cite{background_reason_1,
  background_reason_2, bertero2009image}. With $n, m$ being positive integers, the standard task in such
scenarios is to reconstruct a sparse vector $\bm{x}_{\diamond} \in \mathbb{R}^n$ from the corrupted
signal $\bm{b} \in \mathbb{R}^m$ that follows the Poisson distribution with mean vector $\bm{A}
\bm{x}_{\diamond}$, \textit{i.e.,} $\bm{b} \sim \Poisson (\bm{A}\bm{x}_{\diamond})$, where $\bm{A} \in
\mathbb{R}^{m \times n}$ is a known sensing matrix, and $\Poisson(\cdot)$ acts independently on each
entry of the vector $\bm{A}\bm{x}_{\diamond}$. To estimate $\bm{x}_{\diamond}$, a standard approach is to
cast the problem as the minimization of an objective function that combines a data-fidelity term based on
the celebrated Kullback-Leibler (KL) divergence with a sparsity-inducing regularizer~\cite{spiral_tap,
  essafri2024ell1}.

The KL divergence is a specific instance of the Bregman $D_{\phi}(\cdot, \cdot)$ one, where $\phi$ is the
Boltzmann-Shannon entropy (see Section~\ref{sec:preliminaries} for definitions). Owing to its asymmetry,
two natural candidates arise for the data-fidelity loss: \textbf{(i)} $f(\bm{x}) = D_{\phi}(\bm{b},
\bm{A}\bm{x})$, which corresponds to the statistically natural formulation, and \textbf{(ii)} $f(\bm{x})
= D_{\phi} (\bm{A}\bm{x}, \bm{b})$, which is adopted in~\cite{mirror_descent} for solving inconsistent
linear systems with nonnegative data. While the latter lacks a direct Poisson-likelihood interpretation,
it nonetheless provides a meaningful measure of residuals between nonnegative
signals~\cite{csiszar1991least}. A major challenge in Poisson inverse problems is that, in both cases,
the gradient of $f(\cdot)$ fails to be Lipschitz continuous, thereby preventing standard
proximal-gradient methods from offering convergence guarantees~\cite{beck2009fast}. To overcome this
limitation, the NoLips algorithm~\cite{mirror_descent, bolte2018first} establishes convergence guarantees
by relying on a \textit{Lipschitz-like convexity condition}\/ in place of the standard descent lemma (see
Section~\ref{sec:preliminaries} for details). Algorithms based on the aforementioned data-fidelity
formulations combined with $\ell_1$-norm regularization can be found in~\cite{mirror_descent}.

Notwithstanding, although these algorithms ensure convergence for Poisson inverse problems, the use of
the $\ell_1$-norm---a convex penalty---may induce severe estimation bias by underestimating large
coefficients. To mitigate this well-known issue, various nonconvex penalties have been proposed in the
field of sparse optimization, including the minimax-concave (MC) penalty~\cite{zhang2010nearly}, the
smoothly clipped absolute deviation (SCAD)~\cite{fan2001variable}, and the $\ell_q$-(quasi-)norm for $q
\in (0,1)$~\cite{chartrand2007exact}.

A popular approach for incorporating prior information, such as sparsity, into iterative algorithms is
regularization via operators, commonly referred to as plug-and-play (PnP)~\cite{plug-and-play}, in which
the proximity operator in a forward-backward-type algorithm is replaced by a ``plug-in operator.'' In the
context of Poisson inverse problems,~\cite{hurault2023convergent} employed the PnP strategy within the
NoLips framework, where the plug-in operator is a pre-trained ``black-box'' neural-network
denoiser. While such data-driven approaches have demonstrated promising performance, the design of
mathematically explainable operators for bias reduction in Poisson inverse problems remains a fundamental
challenge.

The external-division operator~\cite{suzuki2026externalPart1,suzuki2026externalPart2}, defined as an affine combination of two
proximity operators with both positive and negative weights, has recently been introduced as a
generalization of the firm-shrinkage operator, which corresponds to the proximity operator of the MC
penalty. The external-division operator provides a flexible framework for bias reduction under the PnP
strategy with convergence guarantees~\cite{plug-and-play}. For instance, the effectiveness of the
external division of two proximity operators associated with the octagonal shrinkage and clustering
algorithm for regression (OSCAR) has been demonstrated in~\cite{suzuki2026externalPart1}, highlighting
the broad applicability of the external-division-operator framework.

This paper introduces a novel method for recovering sparse vectors from linear models corrupted by
Poisson noise. The method integrates an external-division operator into the NoLips framework through the
PnP strategy. The proposed external-division operator is newly constructed from two \textit{Bregman}\/
proximity operators, making it well suited to jointly promote sparsity and reduce estimation bias in
Poisson inverse problems. To elucidate its structure, two complementary reformulations are derived. The
first shows that bias reduction is realized in the primal space of the inverse problem, whereas sparsity
promotion occurs in the dual space. The second reformulation further clarifies the bias-reduction
mechanism by revealing that the operator implicitly incorporates a correction term in the dual
space. Finally, numerical tests corroborate the theoretical and algorithmic developments, demonstrating
that the proposed framework significantly outperforms state-of-the-art KL-based methods on synthetic data
and an image restoration problem.

\section{Preliminaries}\label{sec:preliminaries}

\subsection{Notation and definitions}\label{sec:notation_definitions}

Throughout the paper, let $\mathbb{R}$, $\mathbb{R}_{+}$, $\mathbb{R}_{++}$, and $\mathbb{N}$ denote the
sets of real, nonnegative real, strictly positive real, and nonnegative integer numbers, respectively.
Let $\bm{1}_n$ denote the all-ones vector of dimension $n$.  For any \(\bm{x} \in \mathbb{R}^{n}\), the
\(i\)th entry of \(\bm{x}\) is defined by \({x}_{i}\) or \([\bm{x}]_i\).  The closure and interior of a
subset \(C \subset \mathbb{R}^{n} \) are denoted by $\overline{C}$ and $\Int C$, respectively.

A function $f\colon \mathbb{R}^n \rightarrow (-\infty, +\infty] \coloneqq \mathbb{R} \cup \{+ \infty\}$
  is convex if $f(\alpha \bm{x} + (1-\alpha) \bm{\xi}) \leq \alpha f(\bm{x})+(1-\alpha) f(\bm{\xi})$, for
  any $(\bm{x}, \bm{\xi}, \alpha) \in \mathbb{R}^n \times \mathbb{R}^n \times [0,1]$.  Function $f\colon
  \mathbb{R}^n \rightarrow (-\infty, +\infty]$ is called proper if $\dom f \coloneqq \{\bm{x} \in
    \mathbb{R}^n \mid f(\bm{x}) < +\infty\} \neq \emptyset$.  A function $f\colon \mathbb{R}^n
    \rightarrow (-\infty, +\infty]$ is lower-semicontinuous on $\mathbb{R}^n$ if the level set
      $\lev_{\leq a} f \coloneqq \{\bm{x} \in \mathbb{R}^n \mid f(\bm{x}) \leq a\}$ is closed for any $a
      \in \mathbb{R}$.  Let $\Gamma_0(\mathbb{R}^n)$ denote the set of proper lower-semicontinuous convex
      functions from $\mathbb{R}^n$ to $(-\infty, +\infty]$.  Given a proper function \(f \colon
        \mathbb{R}^{n} \rightarrow (-\infty,+\infty]\), the conjugate function or Fenchel conjugate of
          \(f\) is defined by $f^{*} \colon \mathbb{R}^{n} \rightarrow \mathbb{R} \cup {\{+\infty,
            -\infty\}} \colon \bm{u} \mapsto \operatorname*{sup}_{\bm{x} \in \mathbb{R}^{n}}(\, \langle
          \bm{x}, \bm{u} \rangle_{2} - f(\bm{x})\, )$.

A function $\theta \in \Gamma_0(\mathbb{R}^n)$ is essentially smooth if $\theta$ is differentiable on $\Int \dom \theta$
and $\lim_{ k \to +\infty } \|\nabla \theta (\bm{x}_k)\|_2 = + \infty $ for every sequence
$\{\bm{x}_k\}_{k \in \mathbb{N}} \subset \Int \dom \theta$ converging to a boundary point of $\dom \theta$.  A
function $\theta \in \Gamma_0(\mathbb{R}^n)$ is a Legendre function if $\theta$ is essentially smooth and strictly
convex on $\Int \dom \theta$ \cite{mirror_descent}.  If $\theta$ is a Legendre function, $\nabla \theta$, often referred
to as the \textit{mirror map,} is a bijection from the primal space $\Int \dom \theta$ to the dual space $\Int
\dom \theta^*$.
The Bregman divergence with respect to a given Legendre function $\theta \colon \mathbb{R}^n \rightarrow (-\infty, +\infty]$ is defined as $D_\theta \colon \dom \theta \times \Int \dom \theta \rightarrow \mathbb{R} 
    \colon (\bm{\xi}, \bm{x}) \mapsto \theta(\bm{\xi}) - \theta(\bm{x}) - \langle \nabla \theta (\bm{x}), \bm{\xi} -
    \bm{x} \rangle_2$.

\subsection{The NoLips algorithm}\label{sec:NoLips}
Let $f, g \in \Gamma_0(\mathbb{R}^n)$ such that (s.t.) $f$ is continuously differentiable, but not
necessarily Lipschitz continuous on $\Int \dom f \neq \emptyset$.
Consider the following minimization problem:
\begin{equation}
  \min\nolimits_{\bm{x} \in C} f(\bm{x}) + g(\bm{x}), \label{eq:problem_def}
\end{equation}
where $C$ is a closed convex set with nonempty interior. 
To address the lack of Lipschitz
continuity of $\nabla f$, the NoLips algorithm was introduced in~\cite{mirror_descent}. NoLips replaces
the Euclidean distance of the proximal gradient algorithm with the Bregman divergence $D_h$ for a Legendre function $h$ chosen so that $C = \overline{\dom h}$.
Specifically, the general update rule of the NoLips algorithm is given by:
\begin{equation}
    \bm{x}_{k+1} = \Prox_{\lambda g}^{h} (\, \nabla h^*(\,  \nabla h(\bm{x}_k) - \lambda \nabla f(\bm{x}_k)\,
    )\, ), \label{eq:Nolips_algorithm}
\end{equation}
where $\lambda \in \mathbb{R}_{++}$ is the step size. Here, the \textit{Bregman
  proximity operator}\/ of $g$ of index $\gamma \in \mathbb{R}_{++}$ is defined as
\begin{align}
  \Prox_{\gamma g}^{h} & \colon \Int \dom h \rightarrow \mathbb{R}^n \nonumber \\
  & \colon \bm{z} \mapsto \argmin\nolimits_{\bm{\xi} \in \mathbb{R}^n} \left( g(\bm{\xi}) + \gamma^{-1}
  D_h(\bm{\xi}, \bm{z})\right).
\end{align}
Specifically, when $h \colon \bm{x} \mapsto (1/2)\|\bm{x}\|_2^2$, the Bregman divergence reduces to the
squared Euclidean distance, and the Bregman proximity operator becomes the standard proximity
operator~\cite{bauschke2017convex}
\begin{align}
  \Prox_{\gamma g}\colon \mathbb{R}^n \rightarrow \mathbb{R}^n\colon \bm{z} \mapsto \argmin_{\bm{\xi}
    \in \mathbb{R}^n} \left( g(\bm{\xi}) + \tfrac{1}{2 \gamma}\|\bm{z} -
  \bm{\xi}\|_2^2\right). \label{eq:Prox}
\end{align}
The convergence of the NoLips algorithm is analyzed under a \textit{Lipschitz-like convexity condition,}
stated as:
\begin{align}
  & \exists L \in \mathbb{R}_{++}\ \text{s.t.}\ L h - f\ \text{is convex on}\ \Int \dom  h. \label{eq:LC}
\end{align}

Let now the Boltzmann-Shannon entropy $\phi\colon \mathbb{R}_{+}^m \rightarrow \mathbb{R} \colon \bm{u}
\mapsto \sum_{j=1}^m u_j \log u_j$---known to be of Legendre type---where $0 \log 0 \coloneqq 0$. Under the light
of \eqref{eq:problem_def}, the following two formulations were introduced in \cite{mirror_descent} to
address Poisson inverse problems.
  Note here that $h$ is a Legendre function chosen to ensure $C = \overline{\dom h}$ and to satisfy \eqref{eq:LC}, and it can be distinct from $\phi$ used in the data-fidelity term $f$.
\begin{enumerate}
  \renewcommand{\theenumi}{\normalfont{\textbf{(\roman{enumi})}}}
  \renewcommand{\labelenumi}{\normalfont{\textbf{(\roman{enumi})}}}
\item $f(\bm{x}) \coloneqq D_{\phi}(\bm{b}, \bm{A}\bm{x})$ and $g(\bm{x}) \coloneqq \|\bm{x}\|_1$.  The
  composite loss of \eqref{eq:problem_def} corresponds to Poisson likelihood maximization. 
  In this case, \eqref{eq:LC} is satisfied by choosing Burg's entropy as $h$, \textit{i.e.}, $ h \colon \mathbb{R}_{++}^{n} \rightarrow \mathbb{R} \colon \bm{x} \mapsto -
  \sum_{j=1}^n \log x_j$ \cite[Lemma 7]{mirror_descent}.
\item $f(\bm{x}) \coloneqq D_{\phi}(\bm{A}\bm{x}, \bm{b})$ and $g(\bm{x}) \coloneqq \|\bm{x}\|_1$.  This
  formulation is in general appropriate for solving inconsistent nonnegative linear-system inverse
  problems~\cite{csiszar1991least}. 
  In this case, \eqref{eq:LC} is satisfied by choosing the Boltzmann-Shannon entropy as $h$, \textit{i.e.}, $ h \colon \mathbb{R}_{+}^{n} \rightarrow \mathbb{R} \colon \bm{x} \mapsto \sum_{j=1}^n x_j \log x_j$ \cite[Lemma 8]{mirror_descent}.
\end{enumerate}

\subsection{The external-division operator}\label{sec:ext.div.op}

The external-division operator~\cite{suzuki2026externalPart1, suzuki2026externalPart2} is defined as the
following affine combination of two proximity operators \eqref{eq:Prox}:
\begin{equation}
    \Delta_{\omega} \coloneqq \omega \Prox_{g_1} - (\omega - 1) \Prox_{g_2},
\end{equation}
where $\omega > 1$ and $g_1, g_2 \in \Gamma_0(\mathbb{R}^n)$. A specific example is the firm-shrinkage
operator defined $\forall \gamma, \tau$, with $\gamma > \tau > 0$, as
\begin{align*}
  \firm_{\tau, \gamma} \colon \mathbb{R} \rightarrow \mathbb{R} \colon
  x \mapsto
  \begin{cases}
    x, & |x| \ge \gamma, \\
    \sign(x) \frac{\gamma (|x| - \tau)}{\gamma - \tau} , & |x| \in [\tau, \gamma), \\
      0, & |x| < \tau,
  \end{cases}
\end{align*}
where $\sign(\cdot)$ denotes the standard sign function. Indeed, the firm-shrinkage operator can be
expressed as the external division of two soft-shrinkage operators (\textit{i.e.}, two proximity
operators \eqref{eq:Prox} of the $\ell_1$-norm)~\cite[Proposition 1]{suzuki2026externalPart1}:
\begin{equation}
  \firm_{\tau, \gamma} = \tfrac{\gamma}{\gamma - \tau} \soft_{\tau}- \tfrac{\tau}{\gamma - \tau}
  \soft_{\gamma},
\end{equation}
where, $\soft_{\gamma} \colon \mathbb{R} \rightarrow \mathbb{R} \colon x \mapsto \sign(x) \max \{|x| -
\gamma, 0\}$, $\forall \gamma \in \mathbb{R}_{++}$. The firm-shrinkage operator corresponds to the
identity mapping for inputs of sufficiently large size, and hence induces nearly unbiased
estimates~\cite{zhang2010nearly}.

\section{Main Results}\label{sec:main results}

First, the definition in Section~\ref{sec:ext.div.op} is extended to encompass the external division of
Bregman proximity operators. The resulting operator is then incorporated into the NoLips algorithm via a
plug-and-play (PnP) strategy to address Poisson inverse problems. Two complementary reformulations of the
proposed operator are presented, providing clear structural interpretations. All theoretical results are
stated without proof due to space limitations; complete proofs are deferred to a future publication.

\subsection{External division of two Bregman proximity operators}

To reduce the estimation bias caused by the $\ell_1$-norm, this paper extends the discussion in
Section~\ref{sec:ext.div.op} by defining the external division of two Bregman proximity operators as
follows: for \(\omega, \eta_1, \eta_2, a \in \mathbb{R}_{++}\), with $\omega > 1$ and $\eta_2 > \eta_1$,
and with $h(\cdot)$ being the Boltzmann-Shannon entropy, let
\begin{align}
    T_{\omega, \eta_1, \eta_2, a}^{h} \coloneqq \omega \Prox^{h}_{\eta_1 \lVert \cdot - a \bm{1}_n
      \rVert_1} - (\omega - 1) \Prox^{h}_{\eta_2 \lVert \cdot - a \bm{1}_n
      \rVert_1}. \label{eq:Bregman_external_division_operator}
\end{align}
Here, the shifted $\ell_1$-norm by $a \bm{1}_n$ is employed to induce sparsity around $a$.  It is
empirically shown in Section~\ref{sec:numerical examples} that setting sufficiently small $a > 0$ yields
better reconstruction performance than $a = 0$. The following proposition provides the explicit form of
$T_{\omega, \eta_1, \eta_2, a}^{h}$.

\begin{proposition}
  Let $\eta_{2} \coloneqq \log{(\, (\omega - 1) /(\omega e^{-\eta_1}-1)\, )}$.  Then, for any $\bm{x} \in
  \Int \dom h$, the $i$th entry of $T_{\omega, \eta_1, \eta_2, a}^{h} (\bm{x})$ in
  \eqref{eq:Bregman_external_division_operator}, $\forall i \in \{1, 2, \ldots, n \}$, takes the
  following explicit form:
  \begin{align}
    & [T_{\omega, \eta_1, \eta_2, a}^{h}(\bm{x})]_{i} \nonumber\\
    & = \begin{dcases}
      \left(\omega e^{\eta_1} - (\omega - 1) \kappa \right)
      x_{i}, & x_{i} \in \left[0,a \kappa^{-1}\right), \\
        \omega e^{\eta_1} x_{i} - (\omega - 1) a, & x_{i} \in
        \left[a \kappa^{-1}, a e^{-\eta_1}\right), \\
          a, & x_{i} \in [a e^{-\eta_1}, a e^{\eta_1}], \\
          \omega e^{-\eta_1} x_{i} - (\omega - 1) a, & x_{i} \in \left(a e^{\eta_1}, \,
          a \kappa \right], \\
        x_{i}, & x_i > a \kappa,
    \end{dcases} \label{ext.dev.Bregman.explicit}
  \end{align}
  where $\kappa \coloneqq (\omega - 1) / (\omega e^{-\eta_1}-1)$.
\end{proposition}

\begin{figure}[t!]
    \centering
    \subfloat[]{%
        \includegraphics[width = 0.47\linewidth]{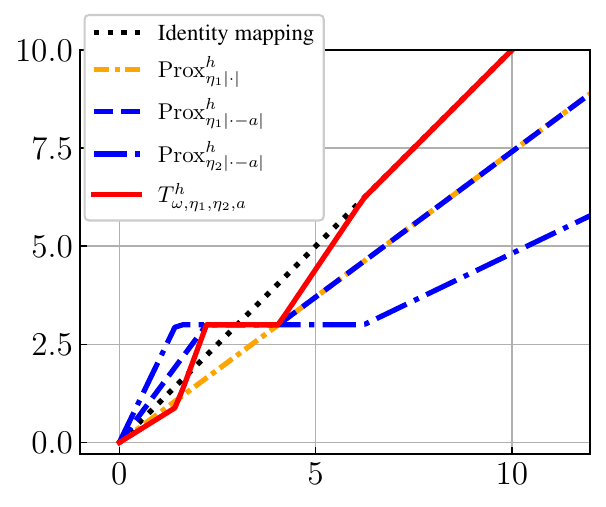}%
        \label{pic:why_external_division}%
    }
    \subfloat[]{%
        \includegraphics[width = 0.47\linewidth]{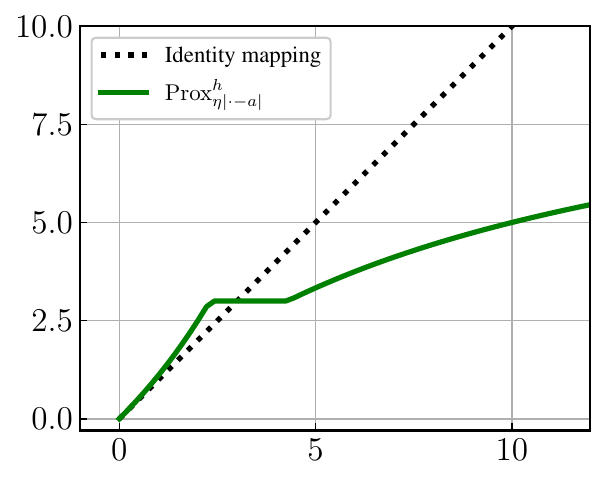}%
        \label{pic:bregman_prox_a}%
    }
    \caption{(a) The difference between the proposed $T_{\omega, \eta_1, \eta_2, a}^{h}$ and the standard
      Bregman proximity operator $\Prox^h_{|\cdot - a|}$, where $h(\cdot)$ is the Boltzmann-Shannon
      entropy, and $\eta_1 = 0.3$, $\omega = 2$, $a=3$. Unlike the proposed $T_{\omega, \eta_1, \eta_2,
        a}^{h}$, the standard Bregman proximity operator leads to estimation bias for sufficiently large
      inputs. (b) The standard Bregman proximity operator $\Prox^h_{\eta |\cdot - a|}$ with Burg's entropy
      $h(\cdot)$ for $a = 3$ and $\eta = 0.1$.}\label{fig:bregman_comparison}
\vspace{-1ex}
\end{figure}

Notice that setting $a = 0$ in \eqref{ext.dev.Bregman.explicit} yields the trivial identity mapping
$T_{\omega, \eta_1, \eta_2, a}^{h} (\bm{x}) = \bm{x}$, $\forall \bm{x} \in \Int \dom h$, and, hence, the
proposed mapping possesses no sparsity-inducing capability. Moreover, whenever $x_i$ is sufficiently
large, \eqref{ext.dev.Bregman.explicit} shows that $[T_{\omega, \eta_1, \eta_2, a}^{h}(\bm{x})]_i =
x_i$. Hence, the proposed operator behaves like the identity one for all sufficiently large inputs,
thereby mitigating estimation bias, in contrast to standard Bregman proximity
operators. Figure~\ref{fig:bregman_comparison} illustrates the marked difference between the proposed
$T_{\omega, \eta_1, \eta_2, a}^{h}$ and the classical Bregman proximity operators.

Algorithm~\ref{alg:proposed} summarizes the proposed framework which employs the PnP
strategy~\cite{plug-and-play}, \textit{i.e.,} the proposed \eqref{eq:Bregman_external_division_operator}
replaces the standard Bregman proximity operator in the NoLips algorithm \eqref{eq:Nolips_algorithm}.

\begin{algorithm}[t!]
  \caption{NoLips with external-division operator}\label{alg:proposed}
  \begin{algorithmic}[1]
    \Statex Arbitrarily fix $\bm{x}_0 \in \mathbb{R}^n$. Choose $\lambda, \eta_1 \in \mathbb{R}_{++}$,
    $\omega > 1$, and $\eta_2 \coloneqq \log (\, (\omega - 1) /(\omega e^{-\eta_1}-1)\, )$.
    Set $f(\cdot) \coloneqq D_{\phi}(\bm{A}\cdot, \bm{b})$.
    \For{$k = 0, 1, 2, \ldots$}
    \State $\bm{x}_{k+1} \coloneqq T_{\omega, \eta_1, \eta_2, a}^{h} (\, \nabla h^* (\, \nabla
    h(\bm{x}_k) - \lambda \nabla f(\bm{x}_k)\, )\,)$.
    \EndFor
  \end{algorithmic}
\end{algorithm}

\begin{remark}
  Algorithm~\ref{alg:proposed} adopts $f(\bm{x}) \coloneqq D_{\phi}(\bm{A}\bm{x}, \bm{b})$, with
  $\phi(\cdot)$ chosen to be the Boltzmann-Shannon entropy, rather than $f(\bm{x}) \coloneqq
  D_{\phi}(\bm{b}, \bm{A}\bm{x})$. These choices are motivated by the following reasons.
  \begin{itemize}
  \item Because $\bm{x}$ is assumed sparse, $\bm{A}\bm{x}$ may contain zero entries. Considering that
    $D_{\phi}$ is defined on $\mathbb{R}^n_{+} \times \mathbb{R}^n_{++}$ for the Boltzmann-Shannon
    entropy $\phi$, $D_{\phi}(\bm{b}, \bm{A}\bm{x})$ is not well defined if zeroes appear in the second
    argument of $D_{\phi}(\cdot, \cdot)$. In contrast, $D_{\phi}(\bm{A}\bm{x}, \bm{b})$ is well defined,
    even if zeroes appear in its first argument, under the convention $0 \log 0 = 0$.
  \item To construct an external-division operator that emulates the identity mapping for large-magnitude
    inputs, and thereby achieves bias reduction, the two constituent Bregman proximity operators in
    \eqref{eq:Bregman_external_division_operator} should exhibit a piecewise linear structure. As shown
    in Figure~\ref{pic:why_external_division}, the operator $\Prox^h_{|\cdot - a|}$ with
    Boltzmann–Shannon entropy $h = \phi$ indeed exhibits a piecewise linear structure, whereas the
    corresponding operator with Burg's entropy $h$, shown in Figure~\ref{pic:bregman_prox_a}, does not.
  \end{itemize}
\end{remark}

\subsection{Two interpretations of the external-division operator}

In this section, two different reformulations of the proposed operator defined in
\eqref{eq:Bregman_external_division_operator} are provided.  First, a decomposition of the Bregman
proximity operator associated with the Boltzmann-Shannon entropy is shown in the following proposition.

\begin{proposition} \label{prop:prox_h_to_prox}
  Let $\tilde{a} \coloneqq \log a + 1$ and $h$ be the Boltzmann-Shannon entropy.
  Then,
  \begin{align}
    \Prox^{h}_{\eta \lVert \cdot - a \bm{1}_{n} \rVert_1}
    = \nabla h^{*} \circ \Prox_{\eta \lVert \cdot - \tilde{a} \bm{1}_{n} \rVert_1} \circ \nabla h.
    \label{eq:prox_h_to_prox}
  \end{align}
\end{proposition}

\begin{corollary}\label{corollary:eq:Bregman_external_division_operator_two_prox}
  Let $\tilde{a} \coloneqq \log a + 1$. Then, the proposed external-division operator
  \eqref{eq:Bregman_external_division_operator} takes the following equivalent form:
  \begin{align}
    T_{\omega, \eta_1, \eta_2, a}^{h}
    {} = {} & \omega \nabla h^{*} \circ \Prox_{\eta_1 \lVert \cdot - \tilde{a} \bm{1}_n \rVert_1} \circ
    \nabla h \notag \\
    & - (\omega - 1) \nabla h^{*} \circ \Prox_{\eta_2 \lVert \cdot - \tilde{a} \bm{1}_n \rVert_1} \circ \nabla h.
  \end{align}
\end{corollary}

Noting that $\nabla h^* = (\nabla h)^{-1}$ and that $\nabla h$ acts as a mirror map,
Corollary~\ref{corollary:eq:Bregman_external_division_operator_two_prox} provides a clear geometric
interpretation of the action of $T_{\omega, \eta_1, \eta_2, a}^{h}(\bm{x})$ on an input vector $\bm{x}
\in \Int \dom h$: \textbf{(i)} $\bm{x}$ is mapped to the dual space via $\nabla h$; \textbf{(ii)} two
shifted soft-shrinkage operators with different thresholds are applied in the dual space; \textbf{(iii)}
the resulting points are mapped back to the original primal space via $(\nabla h)^{-1}$; and
\textbf{(iv)} an affine combination of these two points is computed in the primal space. While sparsity
is induced in the dual space through standard soft-shrinkage operators, bias reduction is achieved in the
primal space via a simple affine combination.

Another interpretation of the mapping \eqref{eq:Bregman_external_division_operator} is provided by the
following proposition. To simplify the discussion, the Boltzmann–Shannon entropy $h$ is assumed to act on
scalars, that is, $h \colon \mathbb{R}_{+} \to \mathbb{R}$. For notational consistency with the preceding
discussion, and with a slight abuse of notation, $\nabla h$ denotes the classical derivative $h'$.

\begin{subequations}
  \begin{proposition} \label{prop:dual_perturbation}
    Let $\tilde{a} \coloneqq \log a + 1$, and
    \begin{align}
      S & \colon \mathbb{R} \rightarrow \mathbb{R} \nonumber \\
      & \colon u \mapsto S_1(u) + \log \left[ \omega \left(\, 1 - \exp\left( S_2(u) - S_1(u) \right)
        \, \right) \right], \label{eq:dual_perturbation_1}
    \end{align}
    where
    \begin{alignat}{2}
      S_1 & \colon \mathbb{R} \rightarrow \mathbb{R} \colon u \mapsto && \Prox_{\eta_1  |\cdot - \tilde{a} |}(u), \\
      S_2 & \colon\mathbb{R} \rightarrow \mathbb{R} \colon u \mapsto && \Prox_{\eta_2  |\cdot - \tilde{a}
        |}(u) + \log (\, (\omega - 1) / \omega\, ).
    \end{alignat}
    Assume that $x_i \ge a\exp(-\eta_1)$. This assumption together with $\eta_2 > \eta_1$ ensures that $1
    - \exp (\, S_2( \nabla h (x_i)) - S_1( \nabla h (x_i))\, ) \in \mathbb{R}_{++}$, so that $S ( \nabla h
    (x_i))$ is well-defined. Then, the following holds true:
    \begin{equation}
      [T_{\omega, \eta_1, \eta_2, a}^{h} (\bm{x})]_i = (\, \nabla h^* \circ S \circ \nabla h\, )
      (x_i). \label{bias.reduction.dual.space}
    \end{equation}
  \end{proposition}
\end{subequations}

Result \eqref{bias.reduction.dual.space} provides insight into the bias-reduction mechanism of
$T_{\omega, \eta_1, \eta_2, a}^{h}$ from the perspective of the dual space. In light of
\eqref{eq:prox_h_to_prox}, the standard one-dimensional Bregman proximity operator can be written as
$\Prox^{h}_{\eta |\cdot-a|} = \nabla h^{*} \circ S_1 \circ \nabla h$, implying that $\nabla h$ is shrunk
by $S_1$ in the dual space. In contrast, $T_{\omega, \eta_1, \eta_2, a}^{h}$ replaces $S_1$ with $S$ by
incorporating the correction term $\log [\omega (1-\exp (S_2(u)-S_1(u)))]$. This correction term
counteracts the shrinkage induced by $S_1$ for an appropriate choice of $\eta_2$. In particular, when
$\eta_{2} \coloneqq \log ((\omega-1)/(\omega e^{-\eta_1}-1))$, it can be shown that $S(u)=u$ for
sufficiently large $u$. This interpretation not only clarifies the bias-reduction mechanism of the
proposed operator in the dual space, but also suggests a possible route toward establishing convergence
guarantees for Algorithm~\ref{alg:proposed} in future work.

\section{Numerical Tests}\label{sec:numerical examples}

\subsection{Synthetic data}

Observed data are modeled as $\bm{b} \sim \operatorname{Poisson}(\bm{A} \bm{x} + \bm{\epsilon})$, where
$\bm{\epsilon} \in \mathbb{R}_{+}^m$ represents background counts, such as dark current in a
charge-coupled-device (CCD) camera \cite{explain_background}. Following \cite{synthetic_experiments}, the
sensing matrix $\bm{A}$ gathers IID entries taking values $0$ or $1/m$ with equal probability.  The
non-zero entries of the sparse ground-truth vector $\bm{x}_{\diamond}$ are drawn uniformly from $[0, k]$
with a given $k \in \mathbb{R}_{++}$ to evaluate performance under high-variance conditions. The
parameter $\rho$ is defined here as $(\text{number of non-zero entries of}\ \bm{x}_{\diamond}) / n$. The
background vector $\bm{\epsilon}$ is set equal to the all-ones vector $\bm{1}_m$.

The proposed method is evaluated against two conventional approaches based on the NoLips
framework~\cite{mirror_descent}: \textbf{(i)} a method that minimizes
$D_{\phi} (\bm{A}\bm{x}, \bm{b}) + \lVert \bm{x} \rVert_1$, referred to as reverse (R-)KL, and
\textbf{(ii)} a method that minimizes $D_{\phi}(\bm{b}, \bm{A}\bm{x}) + \lVert \bm{x} \rVert_1$, referred to
as forward (F-)KL. To ensure numerical stability, a small positive constant is added to the arguments of
the logarithms. Performance is assessed using the normalized mean squared error (NMSE), defined as
$\lVert \hat{\bm{x}} - \bm{x}_{\diamond} \rVert_2 / \lVert \bm{x}_{\diamond} \rVert_2$, where
$\hat{\bm{x}}$ denotes an estimate of $\bm{x}_{\diamond}$. For all methods, hyperparameters are selected
via grid search to achieve the best reconstruction performance.

\begin{figure}[t!]
  \centering
  \subfloat[Learning curves.\label{pic:convergence_iteration}]{
    \includegraphics[width=0.45\linewidth]{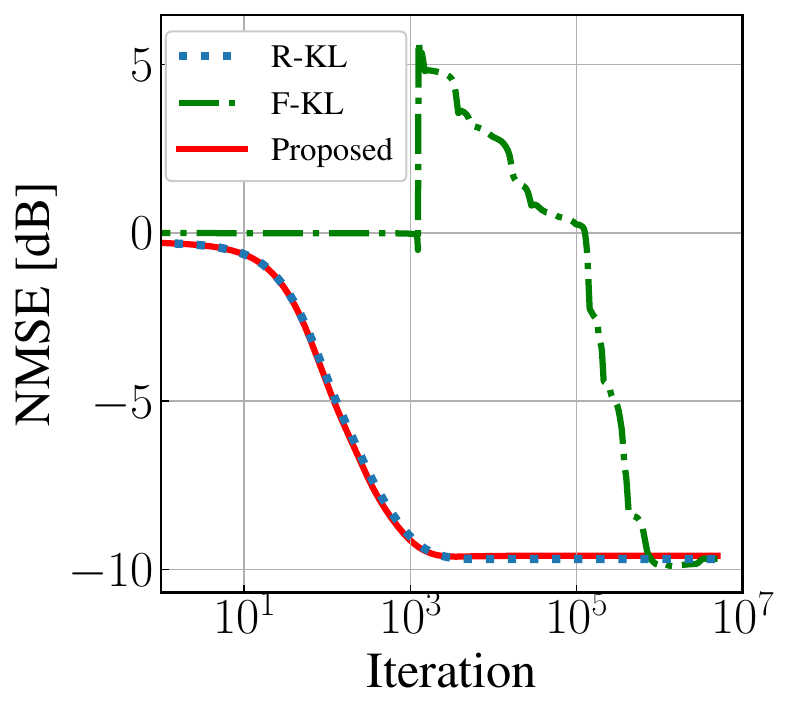} }
  \hfill %
  \subfloat[Distance between successive iterates.\label{pic:distance_iteration}]{
    \includegraphics[width=0.45\linewidth]{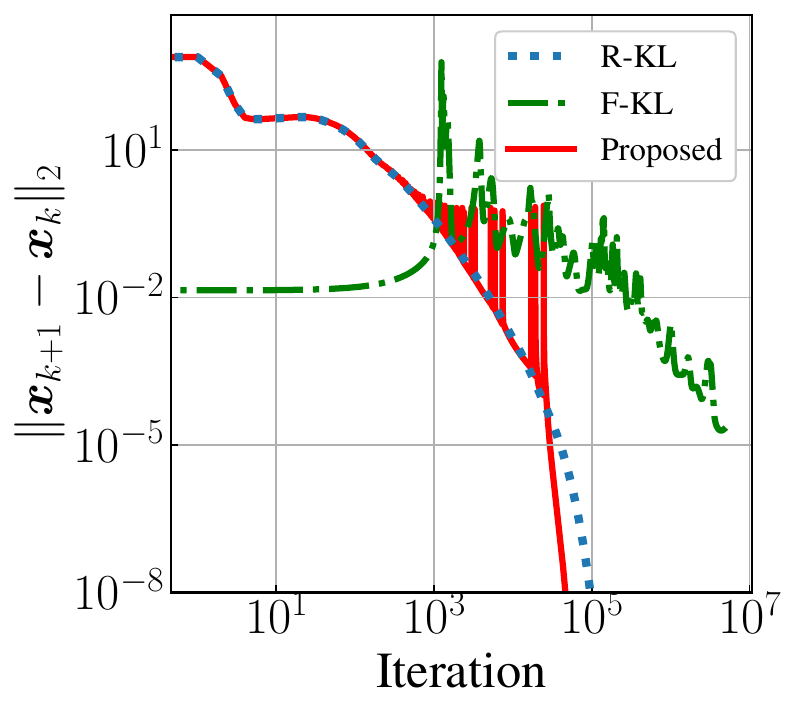} }
  \caption{Convergence behavior for $m=100$, $n=150$, and $\rho=0.1$.}
  \label{pic:convergence_behavior}
  \vspace{-3ex}
\end{figure}

\begin{figure}[t!]
  \centering
  \subfloat[Over-determined case ($m=150$, $n=100$).\label{pic:per_sparsity_100}]{
    \includegraphics[width = 0.45\linewidth]{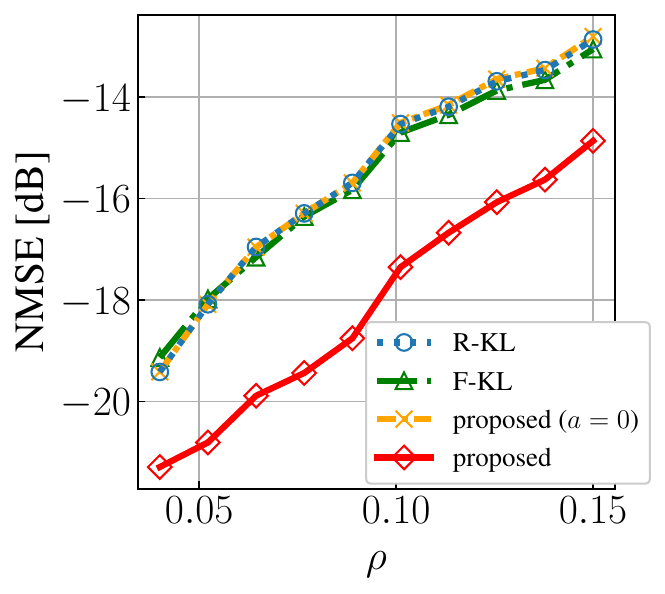} }
  \hfill %
  \subfloat[Under-determined case ($m=100$, $n=150$).\label{pic:per_sparsity}]{
    \includegraphics[width = 0.45\linewidth]{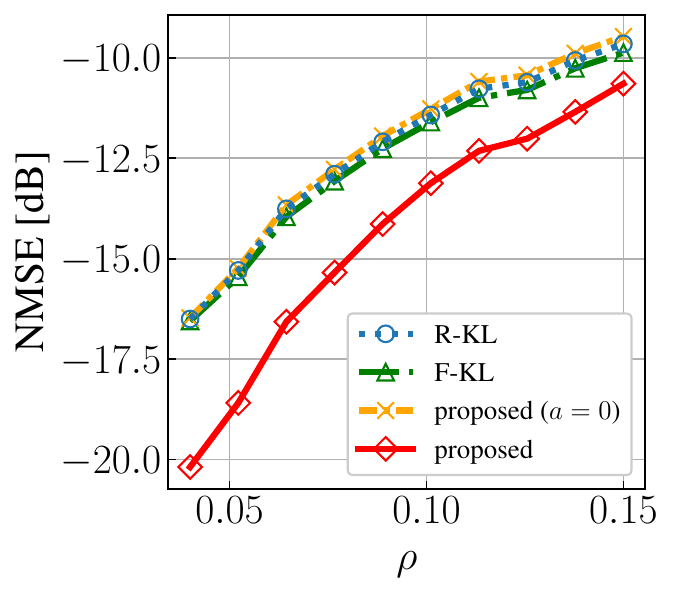} }
  \caption{NMSE as a function of $\rho$ for (a) the over-determined case and (b) the
    under-determined case.}\label{fig:sparsity_ratio}
\vspace{-2ex}
\end{figure}

Figure~\ref{pic:convergence_iteration} illustrates the convergence behavior for $m=100$, $n=150$, and
$\rho=0.1$. The F-KL approach requires approximately $1.0 \times 10^{6}$ iterations to converge, whereas
both R-KL and the proposed method reach a steady state in fewer than $10^{4}$ iterations. This disparity
arises from the step-size selection in the NoLips framework: the upper bound for the F-KL step-size
depends on $1/ \lVert \bm{b} \rVert_1$, which becomes small as $\lVert \bm{x} \rVert_2$ increases. In
contrast, R-KL admits a step size that depends on $\bm{A}$ and remains independent of $\lVert \bm{x}
\rVert_2$, thereby enabling substantially faster convergence. Figure~\ref{pic:distance_iteration}
indicates that F-KL exhibits instability, with \(\lVert \bm{x}_{k+1} - \bm{x}_k \rVert_2\) often
fluctuating above \(10^{-4}\).  In contrast, the NMSE for the proposed method and R-KL reaches a steady
state when \(\lVert \bm{x}_{k+1} - \bm{x}_k \rVert_2\) drops below \(10^{-4}\).

Figure~\ref{fig:sparsity_ratio} reports the average NMSE over $500$ trials for both over- and
under-determined settings. Guided by the previous discussion, the stopping criterion is set to $\lVert
\bm{x}_{k+1} - \bm{x}_k \rVert_2 \le 10^{-4}$ for all methods. To account for the substantial differences
in convergence speed, the maximum number of iterations is set to $10^{4}$ for R-KL and the proposed
method, and to $5 \times 10^{6}$ for F-KL, enabling a fair comparison under realistic computational
budgets. The proposed method consistently achieves much lower NMSE than both F-KL and R-KL in both over-
and under-determined regimes. In addition, the proposed method is compared with its non-translated
counterpart ($a=0$). Since the case $a=0$ reduces to the identity mapping, it fails to induce sparsity
effectively. The pronounced performance gap confirms that the translation operation is essential for bias
reduction and accurate estimation.

\begin{figure}[t]
  \centering
  \subfloat[]{%
    \includegraphics[width=0.32\columnwidth]{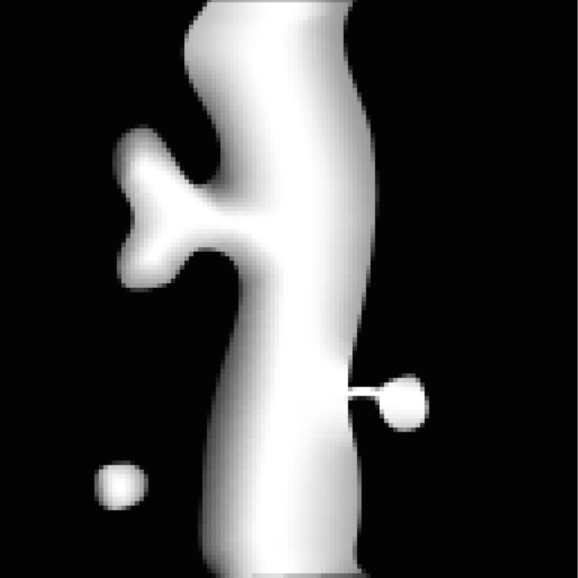}%
    \label{fig:phantom_original}%
  }\hfill
  \subfloat[]{%
    \includegraphics[width=0.32\columnwidth]{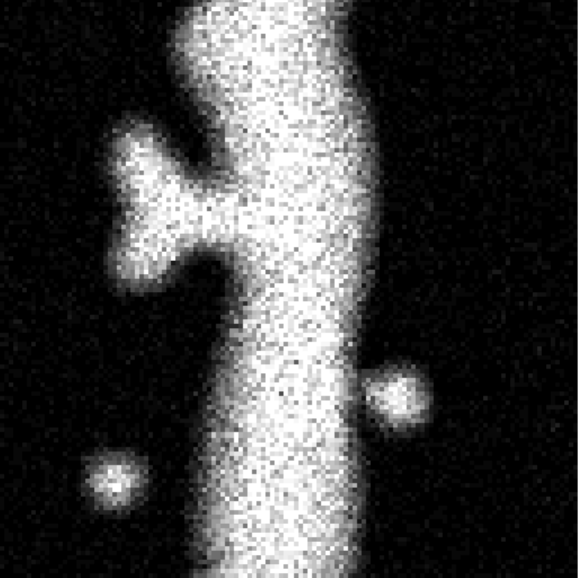}%
    \label{fig:phantom_noisy}%
  }\hfill
  \subfloat[7.55 dB]{%
    \includegraphics[width=0.32\columnwidth]{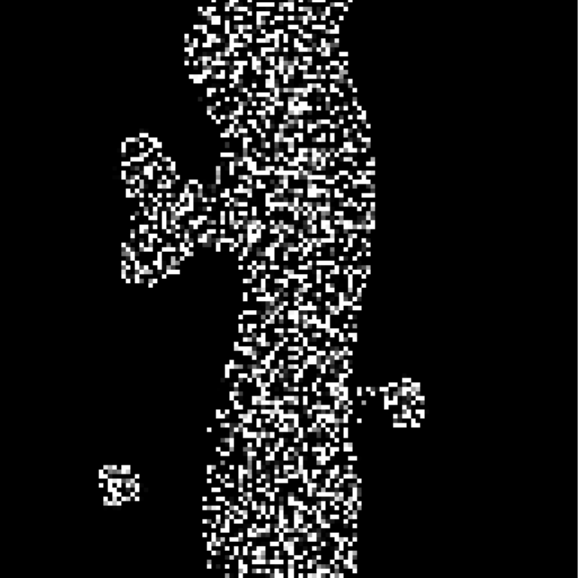}%
    \label{fig:phantom_r_kl}%
  }

  \vspace{1ex}

  \subfloat[13.75 dB]{%
    \includegraphics[width=0.32\columnwidth]{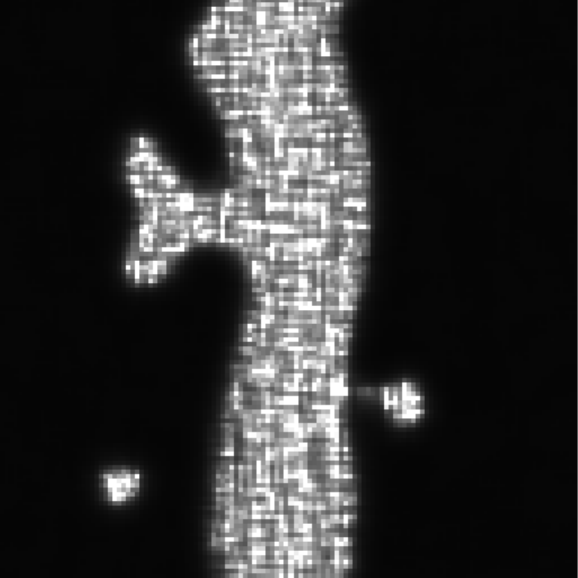}%
  }\hfill
  \subfloat[8.57 dB]{%
    \includegraphics[width=0.32\columnwidth]{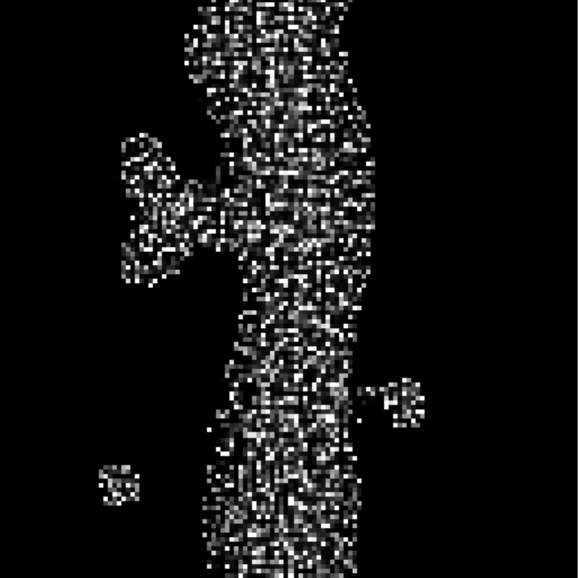}%
  }\hfill
  \subfloat[20.46 dB]{%
    \includegraphics[width=0.32\columnwidth]{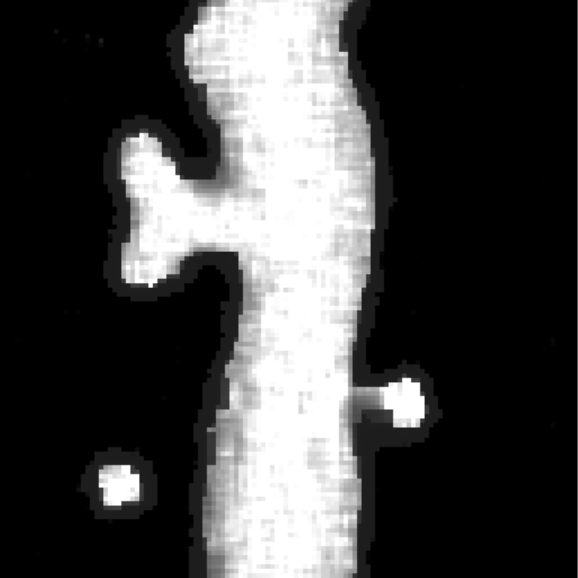}%
    \label{fig:phantom_proposed}%
  }

  \caption{(a) Original, (b) Blurred and noisy, (c) R-KL, (d) F-KL, (e) proposed method ($a=0$), and (f)
    proposed method ($a > 0$).}
  \label{fig:vertical_comparison}
\vspace{-2ex}
\end{figure}

\subsection{Image restoration}\label{sec:application_to_image_denoising}

A phantom image of a neuron with a mushroom-shaped spine~\cite{dupe2009proximal} is used as the
ground-truth image (Figure~\ref{fig:phantom_original}), with $\bm{x}_{\diamond}$ denoting its vectorized
form, where $n$ is the number of pixels. The observed image (Figure~\ref{fig:phantom_noisy}) is generated
by blurring the original image using a $7 \times 7$ moving-average window and subsequently corrupting the
blurred image with Poisson noise. The restored images are shown in
Figures~\ref{fig:phantom_r_kl}--\ref{fig:phantom_proposed}, with intensities normalized so that the
minimum and maximum values are $0$ and $30$, respectively.

Performance is evaluated using the peak signal-to-noise ratio (PSNR), defined as
$\text{PSNR} \coloneqq L^2 / (\, (1/n) \lVert \hat{\bm{x}} - \bm{x}_{\diamond} \rVert_2^2 \, )$, where
$L \coloneqq 30$ denotes the maximum intensity of $\bm{x}_{\diamond}$. Since F-KL suffers from excessively
small step sizes, its step size is carefully tuned to achieve the best possible performance. Even so, the
restored image produced by F-KL exhibits visible artifacts, likely due to slow convergence. The result
obtained by R-KL appears overly sparse and nonsmooth, reflecting an under-estimation of large coefficients.
In contrast, the proposed method produces a relatively smooth image even without total variation (TV)
regularization, effectively preserving both large and intermediate signal intensities. Finally, a
comparison with the non-translated counterpart ($a=0$) highlights that careful tuning of $a$ is crucial
for the proposed operator to achieve effective bias reduction.

\section{Conclusions}

In this study, a novel reconstruction method for Poisson inverse problems was proposed by integrating an
external-division operator into the NoLips framework. The proposed operator was defined through the
external division of two Bregman proximity operators and was designed to coincide with the identity
mapping for large-magnitude inputs, thereby effectively reducing the estimation bias induced by
$\ell_1$-norm regularization. Two reformulations of the proposed operator were presented to provide
structural interpretations. Specifically, the first reformulation revealed the separation between
sparsity-inducing and bias-reducing operations, while the second reformulation elucidated the
bias-reduction mechanism from the perspective of the dual space associated with the Poisson inverse
problem. Numerical tests demonstrated the effectiveness of the proposed method, showing superior
reconstruction accuracy compared to existing approaches on synthetic data and an image restoration
problem. Future work includes a convergence analysis of the proposed method.

\bibliographystyle{IEEEbib_abbrev}
\bibliography{draft_bib}

\end{document}